\documentclass[10pt,twocolumn,letterpaper]{article}

\usepackage{wacv}

\usepackage{amsmath}
\usepackage{graphicx}
\usepackage{caption}
\usepackage{subcaption}
\usepackage{booktabs}
\usepackage{url}
\usepackage{xr-hyper}
\usepackage[accsupp]{axessibility}

\definecolor{wacvblue}{rgb}{0.21,0.49,0.74}
\usepackage[pagebackref,breaklinks,colorlinks,allcolors=wacvblue]{hyperref}

\title{Reference-Guided Data Curation for 3D Medical Vision-Language Pretraining}

\author{
Faidra Patsatzi$^{1,2}$ \qquad
Ashwin Kumar$^{1}$ \qquad
Chong Wang$^{1}$ \\
Curtis P. Langlotz$^{1}$ \qquad
Reinhard Heckel$^{2,\dagger}$ \qquad
Akshay S. Chaudhari$^{1,\dagger}$ \\[1.5ex]
$^{1}$Stanford University \\
$^{2}$Technical University of Munich \\[1.5ex]
{\tt\small faidrap@stanford.edu}
}

\begin{document}
\maketitle{\renewcommand{\thefootnote}{$\dagger$}\footnotetext{Shared co-senior authorship.}}

\begin{abstract}
Training 3D vision-language models (VLMs) on large medical imaging datasets collected across institutions is constrained by data quality. Vast, uncurated pools of image-report pairs are characterized by redundancy, class imbalance, and cross-institutional heterogeneity. We address this challenge of data curation for improving 3D VLM pretraining by proposing an offline, training-free, reference-guided data curation recipe. Our approach leverages a small, high-quality reference set to independently extract quality signals for each modality of a large, uncurated pool and merges them through a rank-aggregation step into a single ranking that guides sample selection. We validate this method on datasets comprising 35k 3D abdominal CT volume-report pairs from two institutions and 58k 2D chest X-ray image-report pairs. On 3D CT, using our reference-guided data curation strategy, our curated model matches the zero-shot disease classification performance of a state-of-the-art baseline (Merlin) while using $3.3\times$ less training data and a reference set of 5k samples. On 2D chest X-rays, our method yields a curated model that matches or exceeds the zero-shot disease classification performance of a model trained on the full pool while using up to $10\times$ less training data and a reference set of only 74 samples. Ablations show that the selection is stable under resampling of the reference, that smaller reference sets still outperform random sampling, and that label rebalancing does not improve performance. To support reproducibility, we release our code (Appendix~\ref{app:code}).

\end{abstract}
\section{Introduction}
\label{sec:intro}
Vision-Language Models (VLMs) are emerging as a promising technology in radiology, capable of performing complex downstream tasks such as findings classification, report generation, and cross-modal retrieval~\cite{Paschali_2025}. High-performing Computed Tomography (CT) VLMs typically require training on large datasets of paired volumes and reports~\cite{blankemeier_kumar2026merlin}, a demand supported by the steady growth of routinely acquired imaging data (with CT exam volumes increasing by 90\% over a 16-year period) coupled with a shortage of radiologists~\cite{CADTH2024}. Considering the growing data pool and large-scale datasets becoming accessible for pretraining~\cite{merlin_abdominal_ct_dataset_2025,Hamamci_2026}, achieving performant 3D VLM pretraining depends on effective data curation that filters redundancy and mitigates class imbalance, rather than on naive scaling.

Here, we consider two data regimes for future 3D VLM development: i) \textit{single-source}, where the available data pool only contains in-distribution volume-report pairs from a single medical institution, and ii) \textit{heterogeneous}, where large pools are formed by aggregating datasets from multiple institutions, introducing distribution shifts caused by varying scanner protocols, patient characteristics, and disease prevalence. In both regimes, the central challenge shifts from acquiring data to identifying a subset that maximizes performance and cross-distribution robustness. This shift motivates the exploration of data selection mechanisms for 3D CT VLM pretraining.

\begin{figure*}[t!]
    \centering
    \includegraphics[trim={0cm 0cm 3.2cm 0cm}, clip, width=0.68\textwidth]{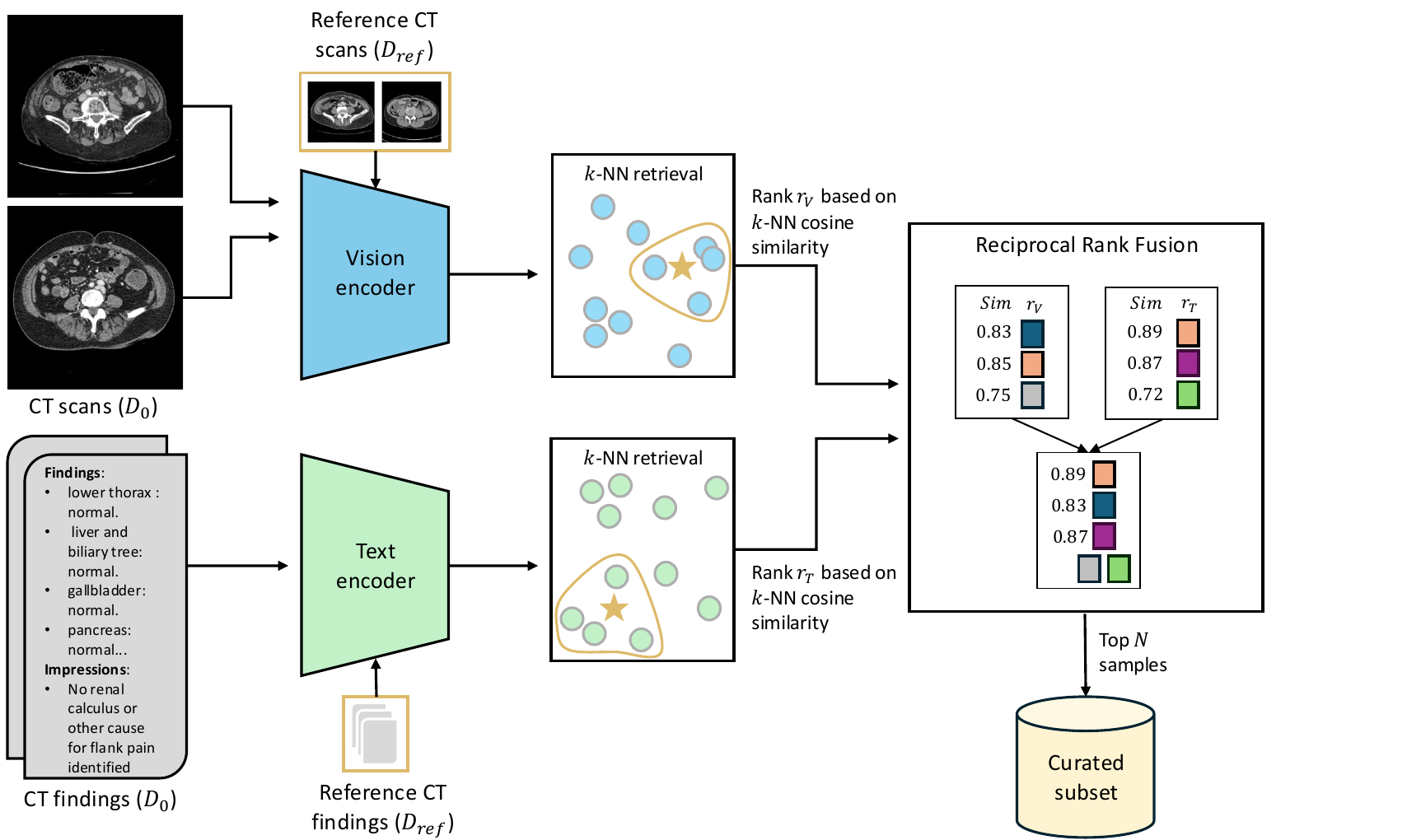}
    \caption{\textbf{Overview of the proposed offline multimodal curation pipeline.} We leverage a small, high-quality reference set ($D_{ref}$) to guide selection from a large uncurated pool ($D_0$). Vision and text streams independently perform $k$-nearest neighbors ($k$-NN) retrieval to produce initial rankings ($r_V, r_T$). These are merged via Reciprocal Rank Fusion (RRF), and the top $N$ samples form the curated subset.}
    \label{fig:method_overview}
\end{figure*}

We propose a reference-guided data curation pipeline. Rather than relying on generic quality metrics or heuristics, we leverage a small, high-quality reference set as an anchor to identify high-quality, reference-aligned samples within large, uncurated hospital data pools. Selecting training data according to its similarity to a curated reference distribution is a well-established principle in foundation-model development in medical and natural domains~\cite{li2025datacomplmsearchgenerationtraining,fang2023datafilteringnetworks,lin2025improvingdeeplearningaccelerated}. However, instantiating this principle for paired 3D volume-report data, in which sample quality has both a visual and a textual component, remains an open question. This is particularly challenging in medicine, where quality judgments require radiological expertise and annotated data are scarce and costly; nearest-neighbor retrieval addresses this by requiring neither labels nor training, ranking an arbitrarily large pool by similarity to a small expert-curated reference set. Our method independently aligns visual and textual features via nearest-neighbor search using open-source embedding models and merges these into a multimodal curated set via rank fusion, thereby matching state-of-the-art 3D VLM pretraining with a fraction $(30\%)$ of the training data (Figure~\ref{fig:method_overview}). At the same data budget, this reference-guided selection outperforms every baseline curation method we evaluate, including random sampling, $k$-center greedy, clustering, CLIP-score filtering, and SemDeDup (Table~\ref{tab:baseline_comparison}). We benchmark reference-guided curation across two distinct medical modalities: 3D Computed Tomography (CT) and 2D Chest X-Ray (CXR). Our key contributions are:

\begin{itemize}
    \item We devise a simple, offline, training-free recipe for multimodal data curation: independent vision and text $k$-NN alignment to a small reference set, merged by reciprocal rank fusion.
    \item On 3D CT, Merlin trained on our curated 4.5k sample subset matches the performance of Merlin trained on the original data distribution, despite using  $3.3\times$ less data, and outperforms all other baseline curation methods at the same budget by 0.09 to 0.21 in distribution (ID) F1 (Table~\ref{tab:baseline_comparison}). Training on a curated subset yields higher cross-institutional F1 both when the second institution contributes to the training pool (0.710 vs.\ 0.689) and when it remains unseen (0.700 vs.\ 0.689), while enriching rare findings and reducing the correlation between per-disease performance and training-set frequency from $r=0.35$ to $r=-0.16$.
    \item The recipe transfers across imaging modalities and architectures (3D CT with an I3D ResNet; 2D chest radiographs (CXR) with a SigLIP ViT) and remains performant across reference set sizes ranging from 5k to 74 samples.
    \item We characterize the robustness of reference-guided curation in our CT experiments. The selection is insensitive to the particular reference draw: performance remains within run-to-run variability when the reference set is resampled. It is moderately sensitive to reference size: reducing the reference set to 200 samples decreases ID F1 by 5.3\%, although the curated subset still exceeds random sampling by 8.5\%. Rebalancing the disease labels of the reference decreases ID F1 by 10.4\% to 14.3\%, and the 2D perceptual curation encoder outperforms native 3D CT foundation models and VLM vision towers by at least 5.8\% ID F1.
\end{itemize}

\section{Related work}
\label{sec:related_work}
Data curation, for example through filtering, is an important component of training foundation models. This is exemplified by
works on data curation for LLMs and vision-language models such as FineWeb~\cite{penedo2024finewebdatasetsdecantingweb}, DataComp \cite{gadre2023datacompsearchgenerationmultimodal}, Data Filtering Networks~\cite{fang2023datafilteringnetworks}, and DataComp-LM~\cite{li2025datacomplmsearchgenerationtraining}, which have demonstrated that effective data curation enables training better models with less data and less compute, or with less data and the same compute. Inspired by this line of research, we investigate how these principles can be adapted to the distinct challenges of image-text paired 3D medical data.

Recent self-supervised learning approaches utilize unsupervised clustering to address data redundancy and long-tail imbalance without external supervision. Vo et al.~\cite{vo2024automaticdatacurationselfsupervised} introduced a hierarchical $k$-means strategy that recursively partitions embeddings to approximate a uniform distribution over the data manifold, effectively upsampling rare concepts. Chen et al.~\cite{chen2025revisitingautomaticdatacuration} adapted this $k$-means clustering method to digital pathology, using hierarchical clusters to capture the multi-scale heterogeneity inherent to whole-slide images and to mitigate class imbalance. However, this hierarchical patch-based method does not directly translate to radiology, with spatially coherent 3D volumes and supervision at the volume–report level. Offline selection heuristics such as coreset construction, embedding deduplication~\cite{abbas2023semdedupdataefficientlearningwebscale}, and image-text alignment scoring form a complementary family of methods; we benchmark against representatives of each in our experiments.

While other existing data curation approaches such as active learning and coreset selection are well established in domains including medical imaging~\cite{zhong2025openpathopensetactivelearning}, methods such as joint example selection~\cite{evans2024datacurationjointexample} and active learning via online clustering~\cite{udandarao2025activedatacurationeffectively} require online updates during training and use very large batch sizes. Given the high memory requirements of 3D VLM pretraining, the applicability of these methods, demonstrated on $2^{15}$ sized batches, is bottlenecked by the availability of specialized hardware to support super-batches of CT volumes.

Parallel to advancements in general language and vision-language models, the medical imaging community is increasingly shifting focus from model architecture to data quality. In 3D medical imaging, Lin et al.~\cite{lin2025improvingdeeplearningaccelerated} show that perceptual-similarity filtering of MRI scans for accelerated MRI matches the gains of a threefold increase in raw data, motivating alignment-based selection for volumetric data.

\begin{figure*}[h!]
    \centering
    \begin{subfigure}{0.59\textwidth}
        \centering
        \includegraphics[width=\textwidth]{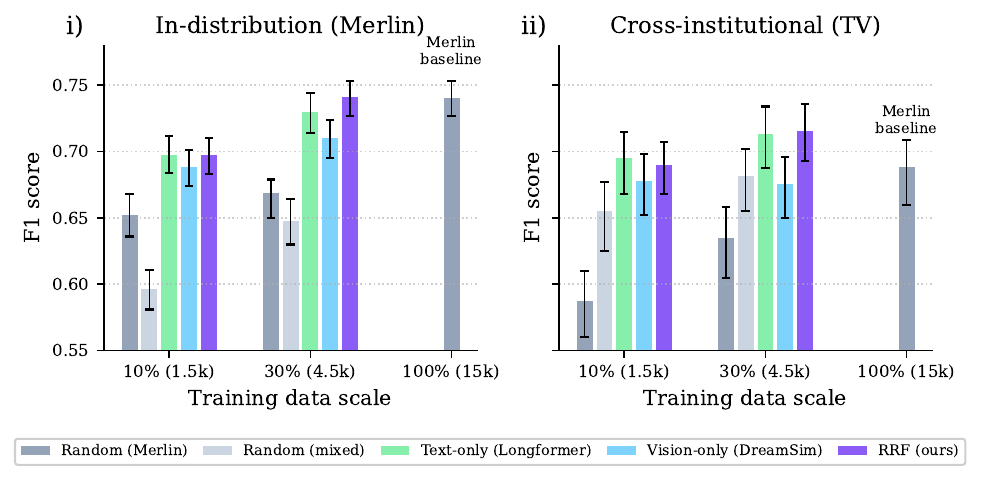}
        \caption{Zero-shot performance per dataset.}
        \label{fig:teaser}
    \end{subfigure}
    \hfill
    \begin{subfigure}{0.40\textwidth}
        \centering
        \includegraphics[width=\textwidth]{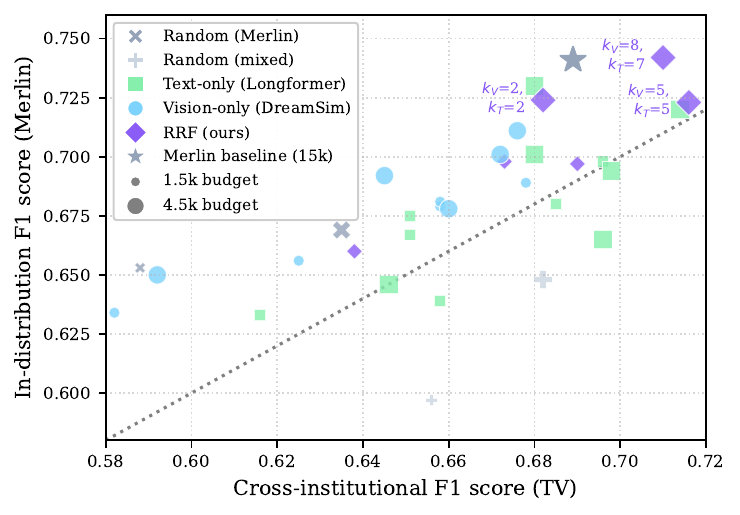}
        \caption{Merlin vs. Tri-Valley (TV) frontier.}
        \label{fig:robustness_tradeoff}
    \end{subfigure}
    \caption{\textbf{Zero-shot findings classification performance.} \textbf{(a)} i) Our RRF curation strategy matches baseline performance (0.741 in F1) using $3.3\times$ less training data. ii) Training on randomly sampled sets from the uncurated dataset or the full Merlin Abdominal CT dataset (Merlin baseline) yields suboptimal cross-institutional performance, reaching 0.635 and 0.689 F1 scores respectively. We plot the best models across $k$ value choices ($k$-NN hyperparameter) for each training data scale and method (text-only, vision-only, RRF). Error bars correspond to 95\% confidence intervals.
    \textbf{(b)} Comparison of Merlin vs. external Tri-Valley (TV) dataset F1 scores. RRF models trace the Pareto frontier.}
\end{figure*}

\section{Problem Formulation}
\label{sec:methods}
We consider a setup where we have an initial, uncurated training data pool $\mathcal{D}_0$, and as mentioned in the introduction, we consider a \textit{single-source} (single institution) regime as well as a \textit{heterogeneous} (multiple institutions) regime.

Our objective is to identify a subset $\mathcal{D}_{C} \subseteq \mathcal{D}_{0}$ that maximizes downstream model performance. We assume we are given a high-quality reference set $\mathcal{D}_{ref}$ (e.g., a curated validation set) that can be used as a proxy for the test set. Consequently, we apply a distribution matching proxy by identifying the subset of $\mathcal{D}_{0}$ that maximizes feature alignment with $\mathcal{D}_{ref}$, following principles grounded in domain adaptation theory~\cite{NIPS2006_b1b0432c}. Distribution matching of this form has strong empirical precedent: classifier-based selection against reference-quality corpora underpins state-of-the-art pretraining data pipelines for language models (DataComp-LM)~\cite{li2025datacomplmsearchgenerationtraining} and for data filtering networks~\cite{fang2023datafilteringnetworks}, and perceptual alignment filtering improves accelerated MRI reconstruction~\cite{lin2025improvingdeeplearningaccelerated}. The open design question is therefore not whether similarity to a reference constitutes an effective selection signal, but how such filtering should be realized for volume-report pairs. We address this question empirically by comparing several realizations of reference-similarity filtering, namely single-modality alignment, early fusion, and rank fusion, against reference-free selection baselines.

\section{Reference-Guided Filtering}
We propose three nearest-neighbor-based matching strategies. By selecting samples from the uncurated pool that lie in the local neighborhoods of the reference samples, we concentrate the training budget on samples with high signal density relative to the reference. Figure~\ref{fig:method_overview} illustrates our approach. We detail our methodology for vision-only, text-only, and vision-text filtering in the subsections below.

\subsection{Vision-Guided Alignment}
\label{sec:vision_alignment}
Following previous work by Lin et al.~\cite{lin2025improvingdeeplearningaccelerated} in alignment-based data filtering for accelerated MRI, we utilize DreamSim~\cite{fu2023dreamsimlearningnewdimensions}, a metric created by ensembling foundation models (CLIP, OpenCLIP, DINO) to capture semantic structure that better aligns with human visual judgment.

Let $\mathcal{X}$ denote the space of 2D images. We define $f_V: \mathcal{X} \to \mathbb{R}^{d_v}$ as the DreamSim encoder mapping an image to a $d_v$-dimensional embedding vector. Let $x, y \in \mathcal{X}$ represent 2D image slices sampled from the uncurated pool $\mathcal{D}_0$ and the reference set $\mathcal{D}_{ref}$, respectively. We denote $\cos(f_V(x), f_V(y))$ as the cosine similarity between their embeddings.

For each reference sample $y \in \mathcal{D}_{ref}$ we iterate over every $x$ in the initial data pool $D_0$ and retrieve the local neighborhood set $\mathcal{N}_y^V$ via $k$-nearest neighbors ($k$-NN):
\begin{equation}
    \mathcal{N}_y^V = \underset{\mathcal{N} \subset \mathcal{D}_0, |\mathcal{N}|=k}{\arg\max} \sum_{x \in \mathcal{N}} \cos(f_V(x), f_V(y))
\end{equation}
\par
The union of these local neighborhoods forms the candidate pool $\mathcal{C}_V = \bigcup_{y \in \mathcal{D}_{ref}} \mathcal{N}_y^V$. To prioritize samples that lie closest to the reference samples, we assign each candidate $i \in \mathcal{C}_V$ a vision alignment score $s_V(i)$ defined by its proximity to the nearest reference sample:
\begin{equation}
    \label{eq:vis_score}
    s_V(i) = \max_{y \in \mathcal{D}_{ref}} \cos(f_V(i), f_V(y)) \quad \text{for } i \in \mathcal{C}_V
\end{equation}
Finally, we construct the curated dataset $\mathcal{D}_V$ of size $N$ by sorting the candidates $i$ by their vision alignment score $s_V(i)$ and taking the top-$N$ samples.

\subsection{Text-Guided Alignment}
Radiology reports pose challenges for standard text encoder models because their length exceeds the token limits of BERT-based models~\cite{devlin2019bertpretrainingdeepbidirectional}, resulting in truncation and loss of critical findings. To address this, we employ Clinical Longformer~\cite{li2022clinicallongformerclinicalbigbirdtransformerslong}, a text encoder that supports a context length of 4,096 tokens to encode the entire findings section of the radiology report. Mirroring the vision pipeline, we select the $k$-nearest neighbors ($k$-NN) in $\mathcal{D}_0$ for each report in $\mathcal{D}_{ref}$ based on embedding similarity.

Let $f_T: \mathcal{T} \to \mathbb{R}^{d_t}$ be the Clinical Longformer encoder. The text alignment process follows the identical protocol defined for vision-guided alignment (Section~\ref{sec:vision_alignment}). Here, $i$ and $y$ represent the textual components (radiology reports) of the samples in $\mathcal{D}_0$ and $\mathcal{D}_{ref}$, respectively.

We compute the text alignment score $s_T(i)$ analogously to Section~\ref{sec:vision_alignment} by calculating the maximum semantic similarity between a candidate report and the reference set:
\begin{equation}
    \label{eq:text_score}
    s_T(i) = \max_{y \in \mathcal{D}_{ref}} \cos(f_T(i), f_T(y)) \quad \text{for } i \in \mathcal{C}_T
\end{equation}
The final text-aligned dataset $\mathcal{D}_T$ consists of the top-$N$ samples sorted by $s_T(i)$.

\subsection{Multimodal Alignment via Reciprocal Rank Fusion}
While single-modality filtering captures specific dimensions of quality, a robust data curation strategy considers both visual and textual quality signals. To combine these effectively, we apply Reciprocal Rank Fusion (RRF)~\cite{36196}.

RRF is a rank aggregation method originally developed to merge results from diverse information retrieval systems. Unlike score-based fusion, which requires normalizing arbitrary score distributions, RRF relies solely on rank positions, making it invariant to the scale of the underlying similarity metrics~\cite{36196}. This property has driven its recent adoption in Retrieval-Augmented Generation (RAG) pipelines~\cite{Rackauckas_2024} and multimodal video search~\cite{samuel2025mmmorrfmultimodalmultilingualmodularized}, where it effectively balances diverse signal sources.

\par In our experiments, early fusion (embedding concatenation before alignment) was unstable across data budgets~(Table~\ref{tab:main_results}), possibly due to the distinct geometric structures of the embedding spaces and limited overlap between vision- and text-aligned sets, as shown in Appendix~\ref{app:venn}. RRF circumvents this by performing late fusion on the candidate lists retrieved by each modality filter individually.

For multimodal alignment, we define the fusion pool as the union of vision and text candidates: $\mathcal{U}_{total} = \mathcal{C}_V \cup \mathcal{C}_T$, and we use the pre-computed similarity scores from the single-modality alignments. Let $s_V(i)$ and $s_T(i)$ denote the similarity scores computed during vision-only and text-only alignment, as defined in previous subsections.

We compute ranks $r_V(i)$ and $r_T(i)$ for all candidates $i \in \mathcal{U}_{total}$ by sorting the candidates in descending order of their similarity scores $s_V$ and $s_T$, assigning rank 1 to the highest score. Samples not present in a specific modality's candidate set are assigned a rank of infinity ($r \to \infty$), effectively penalizing them for absence in that modality's candidate pool. The fused RRF score is then:
\begin{equation}
    \text{Score}_{\text{RRF}}(i) = \frac{1}{\eta + r_V(i)} + \frac{1}{\eta + r_T(i)}
\end{equation}
where $\eta=60$ is a smoothing constant~\cite{36196}.

We select the top-$N$ samples with the highest $\text{Score}_{\text{RRF}}$ to form the final curated dataset $\mathcal{D_\text{RRF}}$. This favors, but does not strictly enforce, samples for which both the vision and the text component align well with the reference set; empirically, 65.8\% of the selected samples lie in the intersection of the vision and text candidate pools (Appendix~\ref{app:venn}). The final vision-language aligned dataset is:
\begin{equation}
\begin{split}
    \mathcal{D}_{\text{RRF}} = \big\{\, x_i \in \mathcal{D}_0 :\ &\text{Score}_{\text{RRF}}(x_i) \\
    &\text{is among the $N$ largest values} \,\big\}
\end{split}
\end{equation}
Since each reference sample contributes at most $k$ neighbors, the candidate union is bounded by $(k_v + k_t)\,|\mathcal{D}_{ref}|$. The $k$ values we experiment with are reported in the results (Section~\ref{sec:results}) and in Appendix~\ref{app:knn}, \ref{app:protocol}.

\section{Experimental Setup}
In this section, we apply reference-guided curation on 3D abdominal Computed Tomography (CT) and 2D Chest X-Ray (CXR) datasets under varying data budgets and measure zero-shot performance. We observe that curated subsets reach similar or better performance than baselines while improving cross-institutional robustness and preserving performance on rare findings.

\subsection{3D Abdominal CT}
We utilize the open-source Merlin Abdominal CT dataset~\cite{merlin_abdominal_ct_dataset_2025} as our in-distribution dataset, comprising 25k pairs of 3D volumes and radiology reports. We use the official validation split of 5k samples, curated and released by the dataset authors~\cite{merlin_abdominal_ct_dataset_2025}, as our reference set $\mathcal{D}_{ref}$. Ablations with reference sets reduced up to $25\times$, resampled reference halves, and a 74-sample CXR reference (Section~\ref{sec:results}) show that the method is not tied to a reference of this size or origin.

We consider two setups: i) a \textit{single-source} data regime consisting of a training pool $\mathcal{D}_0$ of 15k samples (Merlin training set), and ii) a \textit{heterogeneous} data regime with a training pool $\mathcal{D}_0$ of 35k samples consisting of 15k samples from the Merlin training set and 20k samples from a second, institutionally distinct abdominal CT dataset, hereafter referred to as Tri-Valley.

Performance is assessed on a zero-shot classification task based on 30 common abdominal CT findings. We report F1 scores on both the Merlin Abdominal CT test set (5k, in-distribution) and the Tri-Valley test set (5k, cross-institutional). Throughout, in distribution (ID) and out of distribution (OOD) denote evaluation on the Merlin and Tri-Valley test sets, respectively, where OOD is defined relative to the reference distribution; in the heterogeneous regime Tri-Valley samples are present in the training pool, so only the single-source regime evaluates an institution that is entirely unseen during training. The Merlin training split contains 15,331 studies and the validation split 5,023; budgets are expressed relative to the training split, and the 30\% budget corresponds to 4.5k selected studies.

\subsection{2D Chest X-Ray}
We apply our method to 2D radiographic images by considering a \textit{single-source} regime where the training data pool consists of 58k samples from the CheXpert Plus~\cite{chambon2024chexpertplusaugmentinglarge} training set (non-empty findings) and a reference set $\mathcal{D}_{ref}$ of only 74 samples (CheXpert Plus validation set with non-empty findings).

Performance is evaluated via zero-shot classification (12 diseases) on the CheXpert test set (ID, 668 samples) and on the held-out MIMIC-CXR~\cite{PhysioNet-mimic-cxr-2.1.0} test set (OOD, 3.5k samples).

\subsection{Implementation}
For reference-guided filtering, we employ off-the-shelf pretrained encoders to generate the embeddings:
\begin{itemize}
    \item \textbf{Vision encoder:} For 3D CT volumes, we generate embeddings per scan using the DreamSim~\cite{fu2023dreamsimlearningnewdimensions} encoder, which ensembles CLIP, OpenCLIP, and DINO features to capture perceptual similarity (Appendix~\ref{app:protocol}). In Appendix~\ref{app:encoder} we ablate this choice against a native 3D CT foundation model and general-purpose VLM vision towers; DreamSim-based curation outperforms all alternative encoders among the configurations tested, by at least 5.8\% (ID F1). Similarly, for CXR, we use the DreamSim encoder and apply it directly to each CXR image.
    \item \textbf{Text encoder:} For all experiments, we use Clinical Longformer~\cite{li2022clinicallongformerclinicalbigbirdtransformerslong} to embed the findings sections of radiology reports.
\end{itemize}

We train models with the Merlin architecture (Inflated 3D ResNet-152 vision encoder)~\cite{blankemeier_kumar2026merlin} for CT experiments. For CXR experiments, we train a standard ViT-based architecture~\cite{dosovitskiy2021imageworth16x16words} initialized from SigLIP~\cite{zhai2023sigmoidlosslanguageimage}. Full pretraining and evaluation hyperparameters and configurations are also provided in Appendix~\ref{app:reproducibility}.

\subsection{Results for 3D Abdominal CT VLMs}
\label{sec:results}

\begin{table*}[t]
    \centering
    \caption{\textbf{Zero-shot classification results.} Comparison of methods across training data scales and data pool configurations (Mixed: 35k pool of Merlin and Tri-Valley data; Merlin: Merlin Abdominal CT data, 15k) against the Merlin baseline. Text-only, vision-only, vision-text, and RRF model scores correspond to the best-performing model across $k$ value choices ($k$-NN hyperparameter), selected by zero-shot F1 on the Merlin validation split, for each data budget and pool. Budgets are expressed relative to the 15k Merlin training set. For the vision-text method, DreamSim and Clinical Longformer embeddings were concatenated before $k$-NN alignment. }
    \label{tab:main_results}
    \begin{tabular}{lcccc}
        \toprule
        \textbf{Method} & \textbf{Pool} & \textbf{Budget} & \textbf{ID F1} & \textbf{OOD F1}\\
        & & & (Merlin) & (Tri-Valley) \\
        \midrule
        Random sampling & Merlin & 10\% & 0.653 {\scriptsize [0.636, 0.668]} & 0.588 {\scriptsize [0.560, 0.610]} \\
        Text-only (Longformer) & Merlin & 10\% & 0.675 {\scriptsize [0.658, 0.688]} & 0.651 {\scriptsize [0.625, 0.673]} \\
        Vision-only (DreamSim) & Merlin & 10\% & 0.634 {\scriptsize [0.616, 0.647]} & 0.582 {\scriptsize [0.552, 0.608]} \\
        RRF (ours) & Merlin & 10\%  & 0.683 {\scriptsize [0.668, 0.695]} & 0.648 {\scriptsize [0.619, 0.670]} \\
        \midrule
        Random sampling & Mixed & 10\% & 0.597 {\scriptsize [0.581, 0.611]} & 0.656 {\scriptsize [0.625, 0.677]} \\
        Text-only (Longformer) & Mixed & 10\% & \textbf{0.698} {\scriptsize [0.684, 0.712]} & \textbf{0.696} {\scriptsize [0.668, 0.715]} \\
        Vision-only (DreamSim) & Mixed & 10\% & 0.689 {\scriptsize [0.674, 0.701]} & 0.678 {\scriptsize [0.652, 0.698]}  \\
        Vision-text (early fusion) & Mixed & 10\% & 0.633 {\scriptsize [0.616, 0.645]} & 0.616 {\scriptsize [0.587, 0.637]}  \\
        RRF (ours) & Mixed & 10\% & \underline{0.697} {\scriptsize [0.683, 0.708]} & \underline{0.690} {\scriptsize [0.668, 0.707]}  \\
        \midrule[\heavyrulewidth]
        Random sampling & Merlin & 30\% & 0.669 {\scriptsize [0.650, 0.679]} & 0.635 {\scriptsize [0.605, 0.658]}  \\
        Text-only (Longformer) & Merlin & 30\% & \underline{0.730} {\scriptsize [0.714, 0.744]} & 0.680 {\scriptsize [0.652, 0.703]} \\
        Vision-only (DreamSim) & Merlin & 30\% & 0.650 {\scriptsize [0.633, 0.664]} & 0.592 {\scriptsize [0.517, 0.610]} \\
        RRF (ours) & Merlin & 30\% & 0.708 {\scriptsize [0.693, 0.722]} & 0.700 {\scriptsize [0.674, 0.720]} \\
        \midrule
        Random sampling & Mixed & 30\% & 0.648 {\scriptsize [0.630, 0.664]} & 0.682 {\scriptsize [0.655, 0.702]} \\
        Text-only (Longformer) & Mixed & 30\% & 0.701 {\scriptsize [0.685, 0.714]}  & 0.680 {\scriptsize [0.654, 0.701]} \\
        Vision-only (DreamSim) & Mixed & 30\% & 0.701 {\scriptsize [0.685, 0.713]} & 0.672 {\scriptsize [0.642, 0.693]} \\
        Vision-text (early fusion) & Mixed & 30\% & 0.720 {\scriptsize [0.705, 0.732]} & \textbf{0.714} {\scriptsize [0.688, 0.734]}  \\
        RRF (ours) & Mixed & 30\% & \textbf{0.742} {\scriptsize [0.727, 0.753]} & \underline{0.710} {\scriptsize [0.684, 0.729]} \\
        \midrule[\heavyrulewidth]
        Random sampling & Mixed & 100\% & 0.673 {\scriptsize [0.656, 0.686]} & 0.671 {\scriptsize [0.642, 0.692]} \\
        \textit{Merlin baseline} & Merlin &  100\% & 0.741 {\scriptsize [0.727, 0.753]} & 0.689 {\scriptsize [0.660, 0.709]} \\
        \bottomrule
    \end{tabular}
\end{table*}
We first evaluate the hypothesis that reference-guided curation can identify a subset of high-quality samples that enables training 3D VLMs without compromising diagnostic performance. We compare our RRF curation strategy against a range of baselines and single-modality filtering across three data budgets: 10\%, 30\%, and 100\% (15k samples). We train baseline models on subsets retrieved via random sampling, $k$-center greedy selection~\cite{sener2018activelearningconvolutionalneural}, clustering ($k$-means)~\cite{vo2024automaticdatacurationselfsupervised}, CLIP score filtering (computed as the cosine similarity between each scan's CLIP image and report embeddings, following common practice in web-scale filtering~\cite{gadre2023datacompsearchgenerationmultimodal}), and SemDeDup~\cite{abbas2023semdedupdataefficientlearningwebscale}; all baselines operate on the same precomputed embedding spaces and are evaluated at the same data budget (Appendix~\ref{app:protocol}). For reference-guided filtering, we observed low sensitivity to the choice of $k$ in $k$-NN retrieval with the full 5k reference, for which values between 5 and 15 are effective, with our best-performing model using $k=7$ for text and $k=8$ for vision $k$-NN alignment. All $k$ values are selected by zero-shot F1 on the Merlin validation split. To assess training stability, we repeated training with four runs and observed a standard deviation in macro-average F1 of 0.006.

As illustrated in Figure~\ref{fig:teaser} and Table~\ref{tab:main_results}, a model trained on 4.5k pairs selected by RRF from the heterogeneous 35k pool (the 30\% budget) achieves an F1 score of 0.742 on the Merlin test set (ID), matching the performance of the Merlin baseline (0.741). This corresponds to a \textbf{$3.3\times$ reduction in training data} while maintaining high zero-shot disease classification accuracy. Furthermore, RRF outperforms all baseline curation methods at the same data budget (30\%)~(Table~\ref{tab:baseline_comparison}). In contrast, random sampling yields lower performance, consistent with the uncurated pool containing many redundant or low-quality samples. Notably, random sampling from the mixed 35k pool at the full 15k budget reaches only 0.673 ID / 0.671 OOD F1 (Table~\ref{tab:main_results}); RRF at the 4.5k budget therefore surpasses random selection on the same pool by 10.3\% ID F1 and 5.8\% OOD F1 while using less data, indicating that the gain stems from the selection strategy rather than from access to the larger pool. These comparisons additionally provide empirical justification for rank fusion among possible realizations of reference-similarity filtering. Single-modality and early-fusion variants each degrade substantially in at least one setting (e.g., vision-only reaches 0.592 OOD F1 on the Merlin pool at 30\%, text-only 0.701 ID F1 on the mixed pool at 30\%, and early fusion 0.633 ID F1 at 10\%), whereas RRF is the best or within confidence intervals of the best method in every pool and budget~(Table~\ref{tab:main_results}). This pattern reinforces that medical data quality is inherently multimodal and requires considering the quality of both vision and text components.

Beyond zero-shot evaluation, a linear probe on the frozen curated vision encoder attains a mean AUROC of 0.824 on the Merlin disease-classification task, 1.5\% and 5.8\% above random sampling from the Merlin and mixed pools at the same budget, respectively (Appendix~\ref{app:linear_probe}), indicating that the benefits of curation are not specific to the zero-shot protocol.

\begin{table}[t]
    \centering
    \caption{\textbf{Benchmarking against baseline methods.} We benchmark our best-performing RRF model against a range of baseline curation methods on the 4.5k (30\%) data scale applied to the mixed 35k data pool. }
    \label{tab:baseline_comparison}
    \resizebox{\linewidth}{!}{%
    \begin{tabular}{lcc}
        \toprule
        \textbf{Method} & \textbf{ID F1} (Merlin) & \textbf{OOD F1} (Tri-Valley)\\
        \midrule
        Random sampling & 0.648 {\scriptsize [0.630, 0.664]} & 0.682 {\scriptsize [0.655, 0.702]} \\
        K-center greedy~\cite{sener2018activelearningconvolutionalneural} & 0.529 {\scriptsize [0.510, 0.546]} & 0.645 {\scriptsize [0.616, 0.668]}\\
        Clustering~\cite{vo2024automaticdatacurationselfsupervised} & 0.605 {\scriptsize [0.586, 0.620]} & 0.652 {\scriptsize [0.623, 0.675]} \\
        CLIP score-based~\cite{gadre2023datacompsearchgenerationmultimodal} & 0.622 {\scriptsize [0.605, 0.636]} & 0.664 {\scriptsize [0.637, 0.682]} \\
        SemDeDup~\cite{abbas2023semdedupdataefficientlearningwebscale} & 0.631 {\scriptsize [0.615, 0.646]} & 0.633 {\scriptsize [0.604, 0.654]} \\
        \midrule
        RRF (ours) & \textbf{0.742} {\scriptsize [0.727, 0.753]} & \textbf{0.710} {\scriptsize [0.684, 0.729]}\\
        \bottomrule
    \end{tabular}}
\end{table}

\subsubsection{Cross-Institutional Performance}
\label{sec:generalization}
A persistent issue in deep learning for medical imaging is the degradation of model performance on OOD data. In the \textit{heterogeneous}
data regime, we apply our reference-guided curation to the mixed pool using the Merlin validation set as a reference set. The curated subset then includes samples from both institutions. Because Tri-Valley contributes to the training pool in this regime, the comparison against the Merlin-only baseline measures the value of reference-guided inclusion of cross-institutional data rather than generalization to an unseen institution. Figure~\ref{fig:robustness_tradeoff} illustrates the
 resulting trade-off. While the 100\% Merlin baseline reaches a Tri-Valley F1 score of 0.689, our RRF-curated model attains 0.710 while matching Merlin baseline in distribution. By construction, the selection retains the cross-institutional samples that lie closest to the reference in embedding space (Appendix~\ref{app:manifold}), and the observed performance is consistent with this reference-aligned inclusion of cross-institutional data.

The \textit{single-source} data regime provides the unseen-institution comparison: here, both the curated model and the baseline are trained exclusively on Merlin data, and Tri-Valley is entirely unseen. The RRF-curated model reaches 0.700 F1 on the Tri-Valley test set, exceeding the full-data baseline (0.689) by 1.6\% while using 30\% of its training data (Table~\ref{tab:main_results}). This gain suggests that in-distribution data curation might also benefit cross-institutional performance.

\subsubsection{Long-Tail Enrichment and Performance}
Clinical datasets typically exhibit a long-tail distribution, where common findings dominate and rare yet critical pathologies are infrequent~\cite{9363915}. We investigate how reference-guided curation affects this imbalance and whether performance on rare findings is preserved under the reduced training budget.

The curation process consistently results in an enrichment of rare pathologies across three independently curated RRF subsets (Appendix~\ref{app:raredisease}). For example, the relative prevalence of \textit{Renal Hypodensities} and \textit{Hepatomegaly}, two clinically important but infrequent findings, increases considerably in the curated subset compared to the raw pool. This suggests that curation can implicitly undersample the large number of ``normal'' or ``unremarkable'' scans that constitute the majority of routine hospital data and several of the most common findings (e.g., \textit{Hepatic Steatosis}, \textit{Ascites}), which decrease consistently across the three curated subsets.

\begin{figure*}
    \centering
    \includegraphics[width=0.6\textwidth]{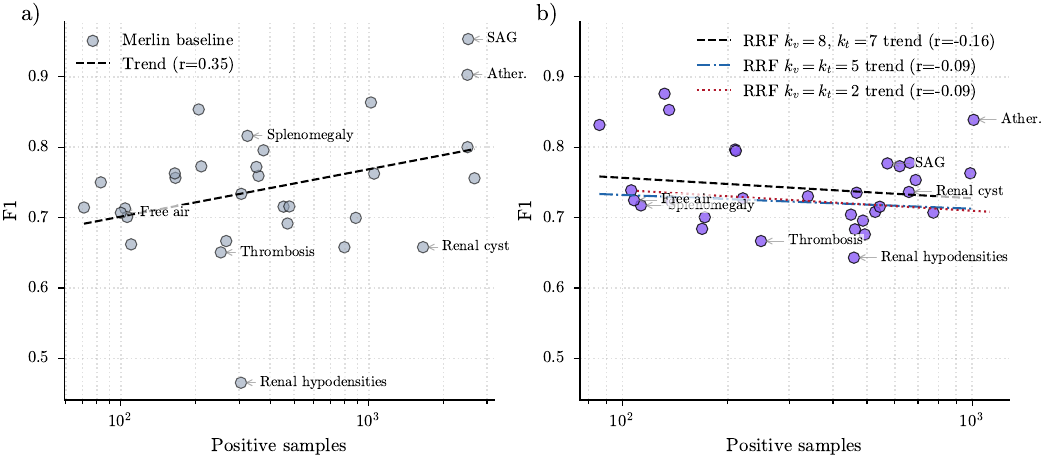}
    \caption{\textbf{Vision-language data curation weakens dependence on training-set prevalence while preserving rare-finding performance.}
    Correlation between positive training sample count and per-disease zero-shot F1. \textbf{(a)} The Merlin baseline shows a positive association with training-set frequency (Pearson $r=0.35$). \textbf{(b)} For the best-performing RRF model ($k_v{=}8$, $k_t{=}7$; markers), the association is much weaker ($r=-0.16$); trend lines for two additional independently curated RRF models ($k_v{=}k_t{=}2$ and $k_v{=}k_t{=}5$) show similar, near-zero association ($r=-0.09$ for both).}
    \label{fig:bias_analysis}
\end{figure*}

The impact of this enrichment is reflected in the per-class performance metrics (Figure~\ref{fig:bias_analysis}). The Merlin baseline model presents a positive correlation ($r=0.35$) between disease prevalence in the training set and classification performance, indicating bias towards majority classes, whereas for the RRF-curated model, this association is weakened ($r=-0.16$): per-disease performance is not correlated with training-set frequency, and performance on rare, tail-end findings is preserved despite the smaller training set. The same near-zero association holds for independently curated and trained RRF models with $k{=}2$ and $k{=}5$ ($r=-0.09$ for both). We note that these correlations are computed over 30 findings and we regard this analysis as descriptive. Details on training stability and variance are provided in Appendix~\ref{app:train_variance}.

\subsubsection{Reference set}
\label{main:ref_set}
\paragraph{Reference set size.} The reference set used in our CT experiments consists of 5k samples (Merlin validation split). To investigate the effect of decreasing the reference set size, we experiment with reference sets of sizes 200 and 1000, sampled from the Merlin validation set, scaling $k$ so that the candidate union exceeds the selection budget. Even with a $25\times$ smaller reference of 200 samples, RRF curation outperforms random sampling from the same pool by 8.5\% ID F1 and 4.8\% OOD F1, and with 1000 samples by 9.3\% ID F1 (Table~\ref{tab:ref_set_size}), indicating that reference-guided curation with a few hundred reference samples effectively guides selection and yields performance gains over random sampling.
\begin{table}[h]
\centering
\caption{\textbf{Reference set size ablation.} Random sampling is denoted as "RS" in the method column.}
\label{tab:ref_set_size}
\resizebox{\linewidth}{!}{%
\begin{tabular}{lcccc}
\toprule
\textbf{Ref. set size} & \textbf{Method} & \textbf{Pool} & \textbf{ID F1} (Merlin) & \textbf{OOD F1} (Tri-Valley)\\
\midrule
- & RS & Merlin & 0.669 {\scriptsize[0.650, 0.679]} & 0.635 {\scriptsize[0.605, 0.658]} \\
- & RS & Mixed & 0.648 {\scriptsize[0.630, 0.664]} & 0.682 {\scriptsize[0.655, 0.702]} \\
200 & RRF (k=50) & Mixed & 0.692 {\scriptsize[0.676, 0.707]} & \textbf{0.715} {\scriptsize[0.691, 0.735]} \\
200 & RRF (k=100) & Mixed & \underline{0.703} {\scriptsize[0.687, 0.716]} & 0.681 {\scriptsize[0.654, 0.701]} \\
1000 & RRF (k=10) & Mixed & 0.679 {\scriptsize[0.661, 0.692]} & 0.632 {\scriptsize[0.603, 0.655]} \\
1000 & RRF (k=20) & Mixed & \textbf{0.708} {\scriptsize[0.692, 0.721]} & \underline{0.698} {\scriptsize[0.673, 0.717]} \\
1000 & RRF (k=40) & Mixed & 0.701 {\scriptsize[0.685, 0.715]} & 0.669 {\scriptsize[0.642, 0.691]} \\
\bottomrule
\end{tabular}}
\end{table}

\paragraph{Reference set choice.} We further test whether curation depends on the specific samples chosen for the reference rather than on their distribution. Two disjoint prevalence-matched halves of the reference select substantially different scans, sharing only 54\% of the curated subset, yet train to F1 within run-to-run noise of each other, indicating that diverse, reference-consistent draws perform similarly well (Appendix~\ref{app:ref_resample}). Rebalancing the reference set's disease prevalence lowers ID F1 by 10.4\% to 14.3\% and the mean F1 of the ten rarest findings by 8.2\% to 9.6\% (Appendix~\ref{app:balanced_ref}). We hypothesize that retrieval by embedding similarity does not necessarily propagate label balance to the curated subset.

\subsection{Results for 2D CXR VLMs}
\begin{figure}
    \centering
    \includegraphics[width=\linewidth]{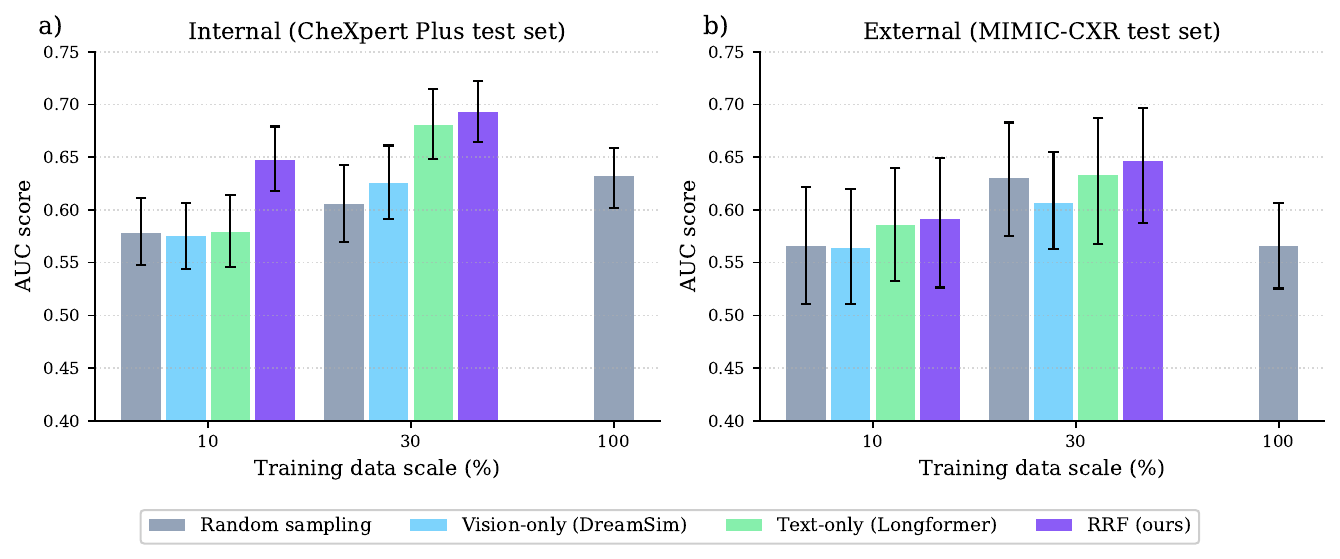}
    \caption{\textbf{Reference-guided data curation for CXR vision-language models.}
    Zero-shot AUC performance on the internal (a) and external (b) test sets. Our curation strategy consistently outperforms uni-modal baselines and random sampling, including the full baseline model (100\%) on the ID test set (CheXpert Plus), while achieving equal or better OOD performance.}
    \label{fig:xray_results}
\end{figure}

Finally, to assess the generality of our approach, we apply our reference-guided curation pipeline to 2D CXR with a reference set of only 74 samples. RRF curation improves over the compared methods in distribution, with an 11.8\% AUC gain over random sampling at the 10\% budget, and is comparable or better out of distribution (Figure~\ref{fig:xray_results}). It also outperforms the 100\% baseline by 2.5\% and 9.8\% ID AUC at the 10\% and 30\% budgets, respectively, and by 4.6\% and 14.3\% OOD AUC. Since our comparisons are budget-controlled with a fixed training recipe, they measure the relative effect of curation rather than state-of-the-art CXR performance (Appendix~\ref{app:cxr_table}). The consistency of the improvements across two modalities and architectures (I3D-ResNet for CT, SigLIP ViT for CXR) supports the method's robustness to a small reference set and suggests that the benefits of reference-guided curation are not tied to a specific model design.

\section{Conclusion}
In this work, we introduce a reference-guided data curation pipeline for efficient 3D medical vision-language pretraining that leverages a
small reference set to select volume-report pairs from large, uncurated data pools. Across 3D CT and 2D CXR benchmarks, our multimodal rank-fusion strategy matches or exceeds state-of-the-art performance while using $3.3\times$ fewer training samples, attains higher cross-institutional F1, and enriches rare diseases in the curated subset while preserving performance on them. These findings suggest that data curation can be as decisive as data scale for efficient and generalizable medical VLM development. Limitations of this work include sensitivity to reference set quality, the fact that our cross-institutional evidence rests on a single external institution, and the Tri-Valley out-of-distribution test set being institutional data that cannot be publicly released, which limits third-party reproduction of the OOD results. Our experiments operate at the scale of tens of thousands of volume-report pairs, the largest currently attainable in an academic setting. As institutional and multi-site pools expand toward millions of studies, training on all available data becomes computationally prohibitive, and label-free, training-free selection strategies such as the one proposed here are likely to become a necessary component of medical VLM pretraining. 
{\small
\section*{Acknowledgments}
F.P.\ is supported by EIKON e.V.\ and the German Academic Exchange Service (DAAD). C.L. is supported by the Medical Imaging and Data Resource Center (MIDRC), which is funded by the National Institute of Biomedical Imaging and Bioengineering (NIBIB) under contract 75N92020C00021. R.H. is supported by the Deutsche Forschungsgemeinschaft (DFG, German Research Foundation) - 456465471,
464123524 and the German Federal Ministry of Education and Research, and the Bavarian
State Ministry for Science and the Arts.
A.C. receives research support from NIH grants R01 HL167974, R01HL169345, R01 AR077604, R01 EB002524, R01 AR079431, P41 EB 027060, P50 HD118632; Advanced Research Projects Agency for Health (ARPA-H) Biomedical Data Fabric (AY2AX000045), Chatbot Accuracy and Reliability Evaluation programs (AYSAX0000024), and Agentic AI-Enabled Cardiovascular Care Transformation (AY2AX000218); and the Medical Imaging and Data Resource Center (MIDRC), which is funded by the National Institute of Biomedical Imaging and Bioengineering (NIBIB) under contract 75N92020C00021 and through ARPA-H.

{
    \small
    \bibliographystyle{ieeenat_fullname}
    \bibliography{main}
}

\appendix
\section*{Appendix}

\section{Code Release}
\label{app:code}
We release our code for curation, training, and evaluation to the public at \url{https://github.com/StanfordMIMI/data-curation-3d-medical-vlm}.

\section{Method Details and Analyses}
\subsection{Fusion Overlap Analysis}
\label{app:venn}
Figure~\ref{fig:venn_analysis} analyzes the overlap between the subsets retrieved by early fusion, RRF, and the single-modality alignments, complementing the discussion of fusion strategies in the main paper.
\begin{figure*}[tb]
    \centering
    \begin{subfigure}[b]{0.66\textwidth}
        \centering
        \includegraphics[width=0.49\textwidth]{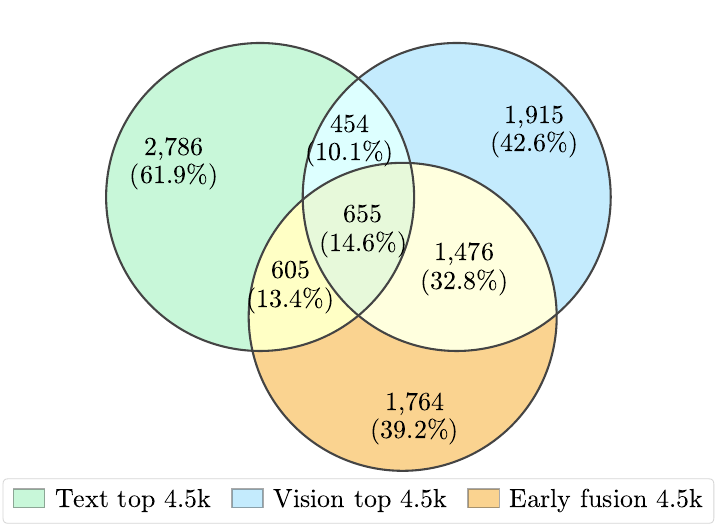}
        \includegraphics[width=0.49\textwidth]{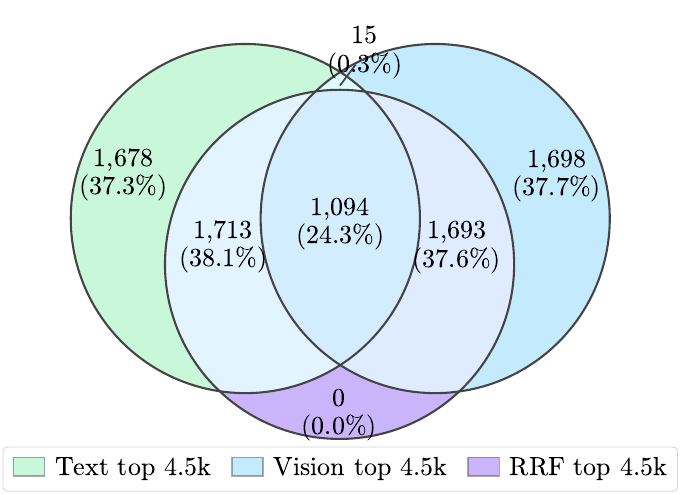}
        \caption{Limitations of early fusion (small intersection)}
        \label{fig:venn_divergence}
    \end{subfigure}
    \hfill
    \begin{subfigure}[b]{0.29\textwidth}
        \centering
        \includegraphics[width=\textwidth]{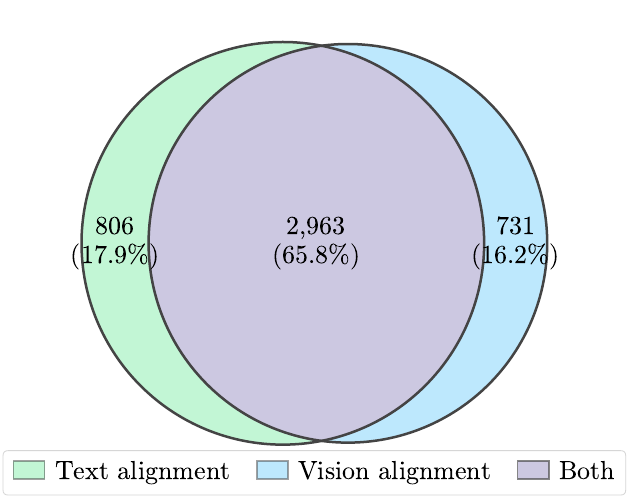}
        \caption{RRF allows for large sample interesection}
        \label{fig:venn_origin}
    \end{subfigure}
    \caption{\textbf{Overlap between RRF, early fusion, and single-modality-aligned sets.}
    \textbf{(a)} Overlap between the top samples ranked by text-guided alignment, vision-guided alignment, early fusion, and RRF strategies. The intersection of only 655 samples between single-modality-aligned sets and early fusion vs. 1,094 samples between single-modality-aligned sets and RRF reveals that RRF retrieves a larger set of top samples from the vision and text-aligned subsets.
    \textbf{(b)} The parent sets of the final RRF-selected subset. Despite the divergence in (a), 65.8\% of samples selected by RRF are drawn from the intersection of the vision and text candidate pools.}
    \label{fig:venn_analysis}
\end{figure*}

\subsection{Manifold Analysis}
\label{app:manifold}
To analyze the selection behavior of RRF, we visualize the vision and text embedding manifolds using UMAP. Figures~\ref{fig:umap}a and \ref{fig:umap}c reveal a distinct domain shift between the Merlin and Tri-Valley datasets. While random sampling from the mixed pool (red) distributes across the entire uncurated pool, RRF curation (blue) acts as a targeted distributional filter. RRF concentrates its selection on manifold regions that best align with the reference set, selecting Tri-Valley samples that are closest to or overlapping with the Merlin data distribution. This selection behavior is consistent with the improved cross-institutional performance of RRF-curated models reported in the main results.

\begin{figure*}[h]
    \centering
    \includegraphics[width=\textwidth]{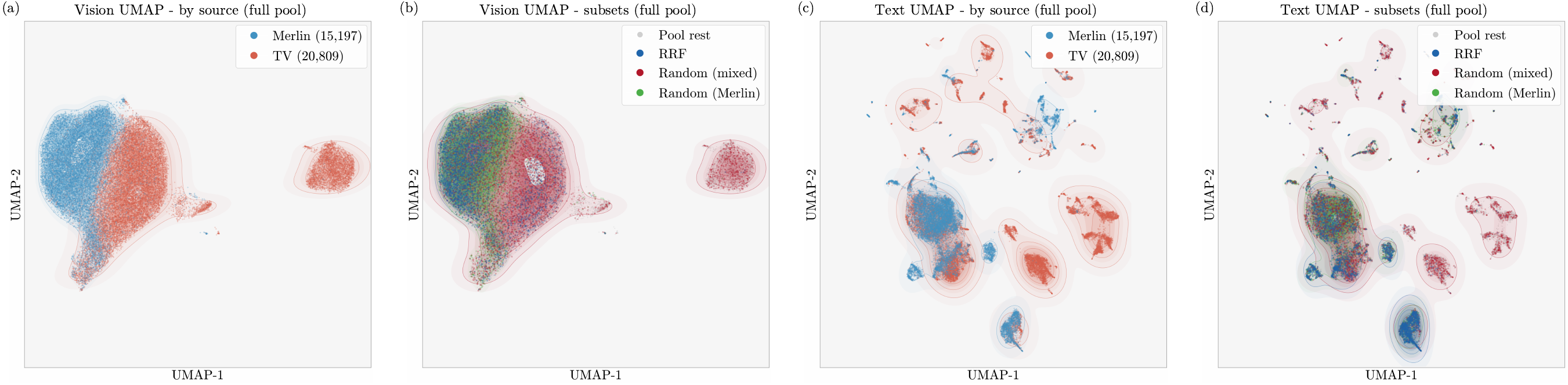}
    \caption{\textbf{UMAP manifold analysis of the heterogeneous 3D CT data pool.} (a, c) Sample distribution by dataset shows clear domain shifts between institutions. (b, d) Compared to the uniform scattering of random sampling, RRF concentrates on regions aligned with the reference set. TV stands for Tri-Valley.}
    \label{fig:umap}
\end{figure*}

\subsection{Choice of Curation Vision Encoder}
\label{app:encoder}
Our curation pipeline is encoder-agnostic: any encoder that yields one embedding per scan can drive the $k$-NN alignment. To test whether the 2D-slice DreamSim signal limits curation quality, we repeat the headline experiment (RRF on the mixed 35k pool, 30\% budget) swapping only the vision embedding used for alignment, with the text encoder, RRF fusion, training, and evaluation unchanged. We compare two encoder families: (i) CT-FM~\cite{pai2025visionfoundationmodelscomputed}, a native 3D CT foundation model pretrained with self-supervision on 148k public CT volumes without reports, and thus free of leakage with respect to our pool; and (ii) the vision towers of two general-purpose 3D VLMs, Qwen3 (Qwen3-VL~\cite{bai2025qwen3vltechnicalreport}) and InternViT (InternVL3.5~\cite{wang2025internvl35advancingopensourcemultimodal}), applied to all axial slices and aggregated either globally or by pooling into six anatomically banded blocks plus a global block (zonal pooling). Table~\ref{tab:encoder_ablation} reports each encoder's best configuration.

No alternative encoder matches DreamSim-based curation: the strongest arms from all three families reach 0.695-0.699 ID F1, roughly 0.04 below DreamSim (0.742), with non-overlapping ID confidence intervals. Zonal pooling improves the globally pooled VLM towers at the paper-standard $k$ configuration. We conclude that neither a native 3D medical encoder nor stronger general-purpose towers improve over DreamSim's perceptual slice similarity for reference-guided curation and that, among the configurations tested, the 2D-slice design is not the primary bottleneck of our method.

\begin{table*}[h]
\centering
\caption{\textbf{Curation vision-encoder ablation.} Zero-shot F1 when the DreamSim slice embeddings used for vision-guided alignment are replaced by alternative encoders (best configuration per encoder and pooling; RRF with Clinical Longformer text, mixed 35k pool, 30\% budget). $k_v$ and $k_t$ denote the vision and text $k$-NN parameters. None of the alternative encoders is trained on the pool.}
\label{tab:encoder_ablation}
\begin{tabular}{llcc}
\toprule
\textbf{Curation vision encoder} & \textbf{Configuration} & \textbf{ID F1} (Merlin) & \textbf{OOD F1} (Tri-Valley)\\
\midrule
DreamSim, 10 axial slices (ours) & $k_v{=}8$, $k_t{=}7$ & \textbf{0.742} {\scriptsize [0.727, 0.753]} & \textbf{0.710} {\scriptsize [0.684, 0.729]} \\
\midrule
CT-FM (native 3D, SSL) & global, $k_v{=}8$, $k_t{=}7$ & 0.698 {\scriptsize [0.681, 0.710]} & 0.686 {\scriptsize [0.661, 0.706]} \\
InternViT (InternVL3.5) & global, $k_v{=}k_t{=}5$ & 0.697 {\scriptsize [0.683, 0.711]} & 0.693 {\scriptsize [0.666, 0.714]} \\
InternViT (InternVL3.5) & zonal pool, $k_v{=}8$, $k_t{=}7$ & 0.688 {\scriptsize [0.672, 0.701]} & 0.687 {\scriptsize [0.663, 0.707]} \\
Qwen3 (Qwen3-VL) & global, $k_v{=}k_t{=}5$ & 0.667 {\scriptsize [0.652, 0.682]} & 0.670 {\scriptsize [0.642, 0.691]} \\
Qwen3 (Qwen3-VL) & zonal pool, $k_v{=}8$, $k_t{=}7$ & \underline{0.699} {\scriptsize [0.683, 0.712]} & \underline{0.699} {\scriptsize [0.675, 0.718]} \\
\bottomrule
\end{tabular}
\end{table*}

\subsection{Curation and Selection Protocol}
\label{app:protocol}
\textbf{Hyperparameter selection.} All $k$-NN parameters ($k_v$, $k_t$) and reported configurations are selected by zero-shot F1 on the Merlin validation split, which also serves as the reference set for curation. The Merlin and Tri-Valley test sets are used exclusively for the final reported evaluations.

\textbf{Candidate-set construction.} The candidate union is bounded by $(k_v + k_t)\,|\mathcal{D}_{ref}|$. In our ablation experiments, $k$ is chosen such that the candidate union exceeds the target budget $N$; when a smaller reference would otherwise not fill the budget, $k$ is increased (e.g., the 200-sample reference ablation uses $k \in \{50, 100\}$, and the prevalence-balanced $T{=}16$ reference uses $k_t{=}15$, $k_v{=}17$). No random fill is used in any reported experiment. Per-experiment $k$ values for CT are given in the corresponding tables.

\textbf{Retrieval implementation.} Retrieval uses exact cosine-similarity search over all pool embeddings; no approximate nearest-neighbor index is used. For 3D CT, we extract ten axial 2D slices (stride 2) from the center of the 3D volume and encode each with DreamSim~\cite{fu2023dreamsimlearningnewdimensions}, which ensembles CLIP, OpenCLIP, and DINO features to capture perceptual similarity; scan-level similarity is computed by averaging the per-slice (slice-aligned) cosine similarities.

\textbf{Baseline implementations.} All curation baselines operate at the same 4.5k budget on the same pool. $k$-center greedy and $k$-means clustering operate on the same multimodal embeddings used for our alignment (DreamSim vision and Clinical Longformer text); SemDeDup operates on the DreamSim vision embeddings; CLIP-score filtering ranks pairs by the cosine similarity between each scan's aggregated CLIP and report embeddings.

\section{Additional Experimental Results}
\subsection{Upsampling of Rare Findings}
\label{app:raredisease}
Figure~\ref{fig:enrichment_histogram} shows the prevalence and F1 score per disease in the subsets retrieved via RRF-based data curation for three independently curated configurations ($k_v{=}8,k_t{=}7$; $k_v{=}k_t{=}5$; $k_v{=}k_t{=}2$) and contrasts them with the prevalence in the Merlin training set (15k) and the F1 score achieved by the Merlin baseline model. The enrichment pattern is nearly identical across the three configurations (22, 22, and 23 of 30 findings upsampled), indicating that it is a property of the curation method rather than of a particular $k$ choice.
\begin{figure*}[h]
    \centering
    \includegraphics[width=\textwidth]{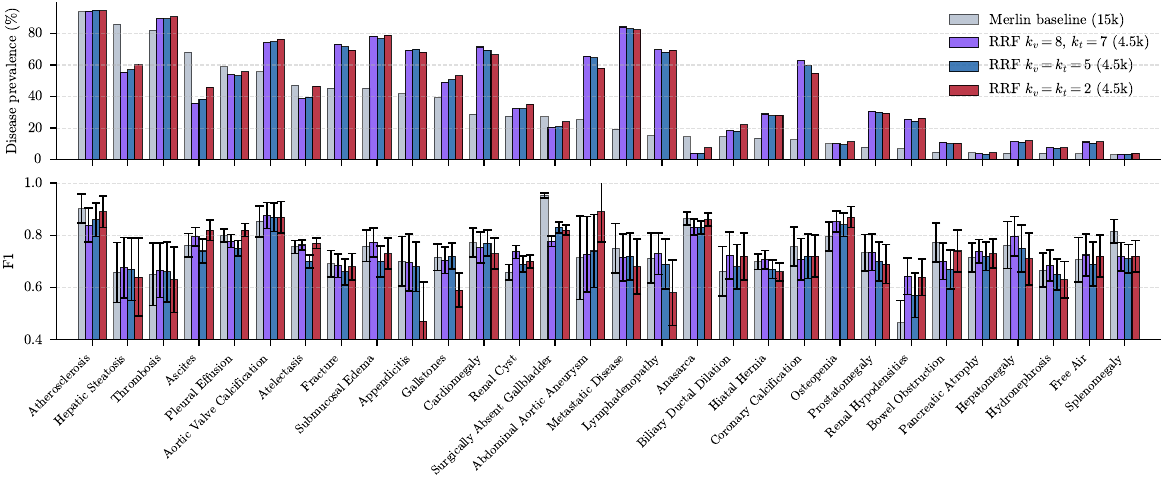}
    \caption{\textbf{RRF functions as a disease prevalence enrichment filter.}
    \textbf{(Top)} Disease prevalence (\%) within the training set of the full Merlin baseline (15k, gray) versus three independently curated RRF subsets (4.5k each; $k_v{=}8,k_t{=}7$ purple, $k_v{=}k_t{=}5$ blue, $k_v{=}k_t{=}2$ red). All three configurations increase the prevalence of most findings, particularly rare, tail-end pathologies such as \textit{Renal Hypodensities} and \textit{Bowel Obstruction}, with nearly identical profiles.
    \textbf{(Bottom)} Corresponding zero-shot F1 scores with 95\% confidence intervals. Across the spectrum of disease rarity, the RRF models remain within the confidence intervals of the baseline for most findings despite being trained on 70\% less data.}
    \label{fig:enrichment_histogram}
\end{figure*}

\subsection{Training Stability}
\label{app:train_variance}
Training stability is supported by a macro-average F1 standard deviation ($\sigma_{\text{macro}}$) of 0.006 across four training runs. Additionally, the mean per-disease standard deviation ($\bar{\sigma}$) of 0.036 is relatively small compared to the primary F1 range of 0.6--0.9. Consistency in variance across the prevalence spectrum, including rare findings such as splenomegaly
($\sigma=0.017$) and common ones such as atherosclerosis ($\sigma=0.025$), suggests that the observed prevalence-performance relationship (main paper, Figure 3) is stable with respect to training randomness; its replication across independently curated RRF configurations ($k_v{=}k_t{=}2$ and $k_v{=}k_t{=}5$, Figure 3 in the main paper and Figure~\ref{fig:enrichment_histogram} in the Appendix) further indicates that it is not specific to a particular curated subset.

\subsection{Sensitivity to Parameter $k$ ($k$-NN)}
\label{app:knn}
We vary $k$ between 4 and 14 for vision- and text-guided alignment at the 1.5k and 4.5k budgets (Figure~\ref{fig:knn_ablation}). With the full 5k reference, performance is only mildly sensitive to $k$ (ID F1 varies by at most $0.046$ across the sweep, with overlapping confidence intervals).
\begin{figure}[h]
    \centering
    \includegraphics[width=\linewidth]{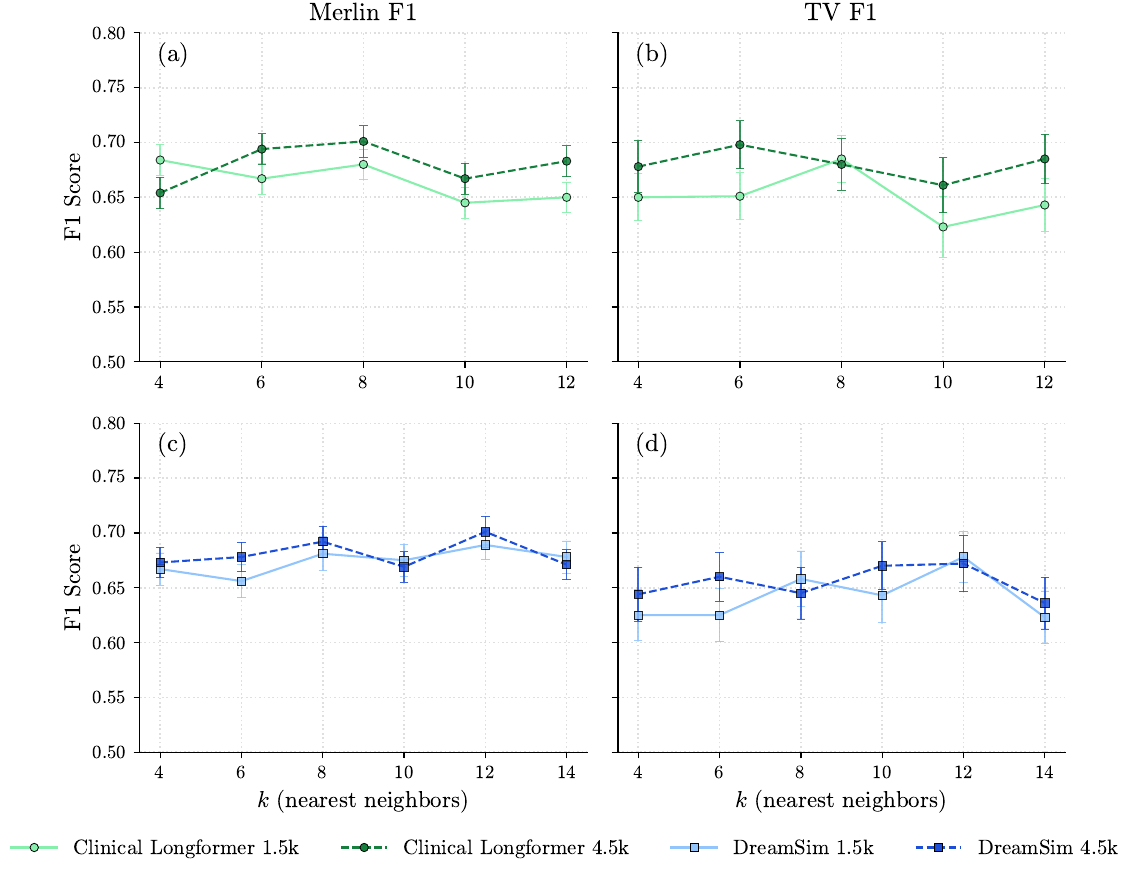}
    \caption{\textbf{Sensitivity to the choice of $k$.} Zero-shot F1 on the Merlin (left) and Tri-Valley (right) test sets for text-guided (top) and vision-guided (bottom) alignment. Error bars are 95\% confidence intervals.}
    \label{fig:knn_ablation}
\end{figure}

\subsection{Reference Set Resampling}
\label{app:ref_resample}
In this ablation experiment, we fix the size and prevalence profile of the reference set and vary only the draw. We split the 5k Merlin validation reference into two disjoint halves (2.5k samples) matched in per-disease prevalence and run the full RRF pipeline ($k_v{=}8$, $k_t{=}7$, mixed 35k pool, 30\% budget) for each half. The two curated subsets differ at the sample level, sharing only 54\% of their 4.5k selections (Jaccard 0.37, Table~\ref{tab:ref_resample}), yet they sample the same region of the pool: their union covers 99.4\% of the full-reference selection, and the Merlin/Tri-Valley composition (roughly 85/15) is unchanged. Despite this turnover, the two models are statistically indistinguishable in test-set performance (F1 0.690 vs.\ 0.694; $\Delta=0.004$, below the run-to-run standard deviation of 0.006, Section~\ref{app:train_variance}); the OOD difference of 0.016 also lies within overlapping confidence intervals. The benefit of reference guidance is therefore not tied to retrieving specific samples; many reference-consistent subsets train equally well. Both halves score about 6.5\% to 7.0\% below the full-reference model, a reference-size effect consistent with the reference set size ablation in the main paper, as $k$ was kept at the headline $k_v{=}8$, $k_t{=}7$ to isolate the draw.

\begin{table}[h]
\centering
\caption{Reference resampling ablation. Two disjoint prevalence-matched halves of the 5k reference, each run through the headline RRF curation (mixed 35k pool, 30\% budget). $J$ denotes Jaccard overlap of the curated 4.5k-sample subsets.}
\label{tab:ref_resample}
\resizebox{\linewidth}{!}{%
\begin{tabular}{lcccc}
\toprule
\textbf{Reference} & $J$(\,$\cdot$\,, other half) & $J$(\,$\cdot$\,, full) & \textbf{ID F1} (Merlin) & \textbf{OOD F1} (Tri-Valley)\\
\midrule
Half A (2512) & 0.369 & 0.575 & 0.690 {\scriptsize [0.675, 0.703]} & 0.667 {\scriptsize [0.639, 0.690]} \\
Half B (2511) & 0.369 & 0.572 & 0.694 {\scriptsize [0.680, 0.706]} & 0.683 {\scriptsize [0.659, 0.701]} \\
Full (5023)   & --    & --    & 0.742 {\scriptsize [0.727, 0.753]} & 0.710 {\scriptsize [0.684, 0.729]} \\
\bottomrule
\end{tabular}}
\end{table}

\subsection{Reference Set Disease Balance}
\label{app:balanced_ref}
The Merlin validation set used as our reference is highly imbalanced across the 30 findings (16 to 903 positives per finding). To test whether this imbalance limits curation, we greedily downsampled it so that every finding retains approximately up to $T$ positives ($T \in \{16, 57, 118\}$, yielding references of 211, 765, and 1326 samples) and repeated RRF curation, training, and evaluation on the mixed 35k pool at the 30\% budget. Balancing via greedy downsampling consistently reduced performance relative to the full reference by 10.4\% to 14.3\% ID F1 (Table~\ref{tab:balanced_ref}), with stronger balancing producing larger drops among the $k$-matched runs. Contrary to the motivating hypothesis, balancing also did not benefit rare findings in distribution: the mean F1 of the ten rarest findings decreases by 9.3\% for $T{=}57$ and 8.0\% for $T{=}118$, with at most three of ten improving. Out of distribution, the rarest findings show a small mean gain of about 4\%, which does not offset declines of 12\% to 27\% among the remaining findings. We hypothesize that retrieval by embedding similarity does not propagate label balance to the curated subset. The three settings carry complementary caveats: $T{=}16$ is the only fully balanced configuration, since findings with fewer than $T$ positives cannot reach the cap for larger $T$, but required a larger $k$ to fill the budget, whereas $T{=}57$ and $T{=}118$ retain the paper-standard $k$ at the cost of partial balance. We further note that balancing necessarily reduces the reference size (from 5,023 to 1,326, 765, and 211 studies), so prevalence and size effects are entangled in this comparison. Reference set size effects are studied in an ablation in Section 5.4.3 of the main paper.

\begin{table}[h]
\centering
\caption{\textbf{Reference set prevalence balancing.} RRF curation from the mixed 35k pool at the 30\% budget, with the reference set greedily downsampled to approximately $T$ positives per finding. $^{\dagger}$$T{=}16$ required $k_t{=}15$ (text) and $k_v{=}17$ (vision) to fill the 30\% budget; all other rows use $k_t{=}7$ and $k_v{=}8$.}
\label{tab:balanced_ref}
\resizebox{\linewidth}{!}{%
\begin{tabular}{lccc}
\toprule
\textbf{Reference set} & \textbf{Size} & \textbf{ID F1} (Merlin) & \textbf{OOD F1} (Tri-Valley)\\
\midrule
Full (unbalanced, ours) & 5023 & \textbf{0.742} {\scriptsize [0.727, 0.753]} & \textbf{0.710} {\scriptsize [0.684, 0.729]} \\
Balanced, $T{=}118$ & 1326 & 0.665 {\scriptsize [0.646, 0.678]} & 0.660 {\scriptsize [0.630, 0.680]} \\
Balanced, $T{=}57$ & 765 & 0.636 {\scriptsize [0.620, 0.649]} & 0.616 {\scriptsize [0.586, 0.640]} \\
Balanced, $T{=}16^{\dagger}$ & 211 & 0.642 {\scriptsize [0.625, 0.657]} & 0.634 {\scriptsize [0.603, 0.657]} \\
\bottomrule
\end{tabular}}
\end{table}

\subsection{CXR Results Table}
\label{app:cxr_table}
Table~\ref{tab:cxr_results} provides the results for CXR experiments.
\begin{table}[h]
\caption{\textbf{Zero-shot classification performance on CXR datasets.}}
\label{tab:cxr_results}
\centering
\resizebox{\linewidth}{!}{%
\begin{tabular}{l c c c}
\toprule
\textbf{Method} & \textbf{Budget} & \textbf{ID AUC} (CheXpert) & \textbf{OOD AUC} (MIMIC) \\
\midrule
Random sampling & 10\% & 0.579 [0.548, 0.612] & 0.567 [0.510, 0.622] \\
Vision-only (DreamSim) & 10\% & 0.576 [0.544, 0.606] & 0.565 [0.511, 0.620] \\
Text-only (Longformer) & 10\% & \underline{0.580} [0.546, 0.615] & \underline{0.587} [0.532, 0.640] \\
RRF (ours) & 10\% & \textbf{0.648} [0.618, 0.679] & \textbf{0.592} [0.526, 0.650] \\
\midrule
Random sampling & 30\% & 0.606 [0.570, 0.643] & 0.631 [0.575, 0.683] \\
Vision-only (DreamSim) & 30\% & 0.626 [0.591, 0.661] & 0.607 [0.563, 0.655] \\
Text-only (Longformer) & 30\% & \underline{0.682} [0.648, 0.715] & \underline{0.634} [0.567, 0.687] \\
RRF (ours) & 30\% & \textbf{0.694} [0.665, 0.722] & \textbf{0.647} [0.587, 0.696] \\
\midrule
\textit{Baseline} & 100\% & 0.632 [0.602, 0.659] & 0.566 [0.526, 0.607] \\
\bottomrule
\end{tabular}}
\end{table}

We note that absolute AUCs in this setting are below those reported for CXR-specific foundation models trained on complete datasets with tuned recipes. Our comparisons are budget-controlled and hold the training recipe fixed (a SigLIP-initialized ViT, which differs in both architecture and training setup from the I3D-ResNet used for CT), therefore measuring the relative effect of curation strategies rather than state-of-the-art CXR performance.

\subsection{Downstream Linear-Probe Evaluation}
\label{app:linear_probe}
To test whether the benefits of curation extend beyond zero-shot classification, we evaluate the pretrained CT image encoders with a linear probe on the Merlin disease-classification task, adopting the linear probe experimental setup and configuration from the Merlin paper~\cite{blankemeier_kumar2026merlin}. We freeze the pretrained I3D-ResNet vision encoder, extract one embedding per volume, and train a linear classification head per finding with a binary cross-entropy objective and the AdamW optimizer, covering the same 30 abdominal findings used in the zero-shot evaluation. We report the AUROC averaged across findings on the Merlin test set, with 95\% confidence intervals from 1000-iteration bootstrap resampling.

As shown in Table~\ref{tab:linear_probe}, the RRF-curated model (best-performing model as highlighted in the main paper) achieves the highest AUROC on the same training budget; this evaluation uses an RRF-curated subset of 4.5k samples from the mixed pool, outperforming random sampling from both the mixed pool and the Merlin pool at the same 4.5k budget. This indicates that the advantages of reference-guided curation persist across both the zero-shot and the linear-probe evaluation protocols.

\begin{table}[h]
\centering
\caption{\textbf{Downstream linear-probe evaluation on the Merlin disease-classification task.} AUROC averaged over the 30 abdominal findings on the Merlin test set, with 95\% bootstrap confidence intervals. All methods use a 4.5k (30\%) training budget.}
\label{tab:linear_probe}
\resizebox{\linewidth}{!}{%
\begin{tabular}{lccc}
\toprule
\textbf{Method} & \textbf{Pool} & \textbf{Budget} & \textbf{AUROC}\\
\midrule
Random sampling & Merlin & 30\% & 0.812 {\scriptsize [0.762, 0.857]} \\
Random sampling & Mixed & 30\% & 0.779 {\scriptsize [0.725, 0.829]} \\
RRF (ours) & Mixed & 30\% & \textbf{0.824} {\scriptsize [0.774, 0.868]} \\
\bottomrule
\end{tabular}%
}
\end{table}

\subsection{Pretraining and Evaluation Details}
\label{app:reproducibility}
\textbf{3D CT.} CT volumes are resampled to $1.5 \times 1.5 \times 3$\,mm spacing, intensity-clipped to $[-1000, 1000]$\,HU, and cropped or padded to $224 \times 224 \times 160$ volumes. We train the Merlin architecture, which pairs an Inflated 3D ResNet-152 vision encoder with a Clinical Longformer text encoder, using a contrastive CLIP objective with a learnable temperature initialized to $\log 10$ and a bias of $-10$. We optimize with AdamW (weight decay $0$) at an initial learning rate of $1\times10^{-5}$ under a cosine learning-rate schedule, using a per-GPU batch size of $18$, mixed-precision (fp16) training, and $100$ epochs.

Zero-shot classification compares each scan embedding against positive and negative prompt ensembles per finding and predicts the class with the higher mean similarity, so no decision thresholds are tuned.  For both zero-shot classification and the linear probe (Section~\ref{app:linear_probe}), our evaluation protocol is identical to that in the Merlin paper~\cite{blankemeier_kumar2026merlin}, including prompt-based scoring and the class-balanced per-finding evaluation.

\textbf{2D CXR.} We train a ViT initialized from SigLIP-Large (patch size 16, pretraining resolution $384 \times 384$), with position embeddings interpolated for $512 \times 512$ inputs, paired with the SigLIP text encoder and optimized with the sigmoid contrastive (SigLIP) objective with a learnable temperature initialized to $0.07$. We use AdamW with a learning rate of $1\times10^{-5}$, weight decay $0.01$, and $1000$ warmup steps, a per-GPU batch size of $64$ on two GPUs, and $10$ epochs.

All reported confidence intervals are 95\% intervals obtained by 1000-iteration bootstrap resampling of the test set. Run-to-run stability is characterized by a macro-averaged F1 standard deviation of $0.006$ across four training runs (Section~\ref{app:train_variance}).

CT models are trained on NVIDIA RTX A6000 GPUs and CXR models on NVIDIA H100 GPUs. All models are evaluated at the final training checkpoint.

\subsection{Compute Cost}
\label{app:compute}
 All CT models are trained for 100 epochs on their respective subsets, so the number of optimization steps scales with subset size. On a single GPU, pretraining on the curated 4.5k subset takes approximately 37 GPU-hours, compared with approximately 140 GPU-hours for the full 15k baseline, a saving of roughly $3.8\times$ per training run. The curation itself adds little: given precomputed embeddings, one alignment configuration (vision and text $k$-NN retrieval plus rank fusion and merging) completes in under one CPU-hour. Embedding the pool with the frozen DreamSim and Clinical Longformer encoders is a one-time cost, below that of a single full-data pretraining run, and is amortized over every curation configuration studied in this work.

\section{Data Governance}
\label{app:governance}
The Merlin Abdominal CT dataset and the CheXpert Plus and MIMIC-CXR datasets are publicly released, de-identified research datasets used under their respective data-use agreements. The Tri-Valley dataset is an institutional, de-identified dataset that cannot be publicly released. We use the official Merlin training, validation, and test splits.
\section{Conflict of Interest Disclosures}
C.L. has received consulting fees from Sixth Street, Everlight Radiology, and Gilmartin Capital and honoraria for lectures from the Singapore Ministry of Health, Philips Medical, Canon Medical, and McKinsey \& Company. He has received travel support from Singapore Ministry of Health, Philips Medical, and McKinsey \& Company.
He has patents pending in collaboration with GE Healthcare.
He serves on the Board of Directors of Bunkerhill Health and Sirona Medical.
He owns equity in Bunkerhill Health, Sirona Medical, Whiterabbit.ai, Galileo CDS, ADRA.ai, Radiology Partners, TurboRadiology, Abxtract, Raidium, and Harrison.ai.
His institution has received licensing fees from Harrison.ai and grants or gifts from AWS, Bayer, BunkerHill Health, Carestream, CARPL.ai, Clairity, GE HealthCare, Google Cloud, IBM, Kheiron, Lambda, Lunit, Microsoft, Nightingale Open Science, Philips, Siemens Healthineers, Stability.ai, Subtle Medical, VinBrain, Visiana, and Whiterabbit.ai.
Unrelated to this work, A.C. receives research support from GE Healthcare, Siemens, Philips, Microsoft, Amazon, Google, NVIDIA, Stability; has provided consulting services to Patient Square Capital, Chondrometrics GmbH, and Elucid Bioimaging; is co-founder of and receives personal fees from Cognita Imaging; has equity interest in Subtle Medical, LVIS Corp, Brain Key, Almanac Health, and Radiology Partners.

\end{document}